\documentclass{article}

\usepackage{arxiv}
\usepackage{array} % 用于自定义列宽居中
\usepackage[utf8]{inputenc} % allow utf-8 input
\usepackage[T1]{fontenc}    % use 8-bit T1 fonts
\usepackage{hyperref}       % hyperlinks
\usepackage{url}            % simple URL typesetting
\usepackage{booktabs}       % professional-quality tables
\usepackage{amsfonts}       % blackboard math symbols
\usepackage{nicefrac}       % compact symbols for 1/2, etc.
\usepackage{microtype}      % microtypography
\usepackage{lipsum}
\usepackage{graphicx}
\graphicspath{ {./images/} }
\usepackage{algorithm}
\usepackage{algorithmic}
\usepackage{float}
\usepackage{booktabs}
\usepackage{amsfonts}
\usepackage{bibentry}
\usepackage{amsmath}

\title{Silent Inconsistency in Data-Parallel Full Fine-Tuning:
Diagnosing Worker-Level Optimization Misalignment}

\author{
 Hong Li \\
  School of Transportation\\
  Southeast University\\
  Nanjing, Jiangsu 211189 \\
  \texttt{hongli@seu.edu.cn} \\
    \And
 Zhen Zhou \\
  School of Transportation\\
  Southeast University\\
  Nanjing, Jiangsu 211189 \\
  \texttt{zzhou602@seu.edu.cn} \\
    \And
 Honggang Zhang\thanks{Corresponding Author} \\
  Department of Logistics and Maritime Studies\\
  The Hong Kong Polytechnic University\\
  Kowloon, Hong Kong, China 999077\\
  \texttt{honggang.zhang@polyu.edu.hk} \\
  \And
 Yuping Luo \\
  School of Transportation\\
  Southeast University\\
  Nanjing, Jiangsu 211189 \\
  \texttt{yp\_luo\_py@163.com} \\
    \And
 Xinyue Wang \\
  School of Transportation\\
  Southeast University\\
  Nanjing, Jiangsu 211189 \\
  \texttt{213241627@seu.edu.cn} \\
    \And
 Han Gong \\
  School of Transportation\\
  Southeast University\\
  Nanjing, Jiangsu 211189 \\
  \texttt{213240417@seu.edu.cn} \\
    \And
 Zhiyuan Liu\footnotemark[1] \\
  School of Transportation\\
  Southeast University\\
  Nanjing, Jiangsu 211189 \\
  \texttt{zhiyuanl@seu.edu.cn} \\
}

\begin{document}
\bibliographystyle{plain}
\maketitle

\begin{abstract}
Data-parallel (DP) training with synchronous all-reduce is a dominant paradigm for full-parameter fine-tuning of large language models (LLMs). While parameter synchronization guarantees numerical equivalence of model weights after each iteration, it does not necessarily imply alignment of worker-level optimization dynamics before gradient aggregation. This paper identifies and studies this latent mismatch, termed \emph{silent inconsistency}, where cross-worker divergence in losses and gradients can remain invisible under conventional aggregated monitoring signals. We propose a lightweight, model-agnostic diagnostic framework that quantifies worker-level consistency using training signals readily available in standard pipelines. Specifically, we introduce three complementary metrics: loss dispersion, gradient-norm dispersion, and gradient-direction consistency measured by inter-worker cosine similarity. The proposed metrics incur negligible overhead and require no modification to model architecture, synchronization mechanisms, or optimization algorithms. We validate the framework by fully fine-tuning the 1B-parameter \texttt{openPangu-Embedded-1B-V1.1} model on the \texttt{tatsu-lab/alpaca} dataset using an 8-NPU DP setup, under controlled perturbations of cross-rank stochasticity. Experimental results show that progressively desynchronized data shuffling and random seeds lead to substantial increases in loss/gradient dispersion and reduced directional alignment, despite smooth globally averaged loss curves. These findings demonstrate that the proposed indicators provide actionable visibility into hidden instability modes in large-scale DP fine-tuning, enabling more reliable diagnosis and configuration assessment.
\end{abstract}

% Uncomment the following to link to your code, datasets, an extended version or similar.
% You must keep this block between (not within) the abstract and the main body of the paper.
% \begin{links}
%     \link{Code}{https://aaai.org/example/code}
%     \link{Datasets}{https://aaai.org/example/datasets}
%     \link{Extended version}{https://aaai.org/example/extended-version}
% \end{links}

\section{Introduction}

Full fine-tuning remains a principal strategy for adapting large language models (LLMs) to downstream tasks, particularly in instruction-following and domain-specialized scenarios. Although parameter-efficient methods such as LoRA~\cite{devalal2018lora} and prefix tuning~\cite{li2021prefix} have been proposed to reduce computational and memory costs, full parameter updates are still widely adopted in practice due to their flexibility and superior task adaptation capability. As model scales expand to billions or even hundreds of billions of parameters~\cite{kaplan2020scaling}, distributed training becomes a computational necessity. Among distributed paradigms, data parallel (DP) training is extensively employed because of its architectural simplicity and scalability across accelerators~\cite{sergeev2018horovod}. In standard DP configurations, each worker computes gradients on local mini-batches and synchronizes them via collective communication, typically all-reduce, thereby ensuring parameter equivalence after each update.

Despite this strict parameter synchronization, equivalence of model parameters does not imply equivalence of optimization trajectories~\cite{zhao2022batch}. Prior to gradient aggregation, each worker independently performs forward and backward propagation on stochastic data shards. These local computations are affected by data ordering randomness~\cite{hayes2025inexact}, gradient stochasticity~\cite{riabinin2025gluon}, mixed-precision arithmetic~\cite{yu2025exploring}, floating-point non-associativity~\cite{perez2023training}, and hardware-level runtime variability~\cite{li2024large}. While all-reduce guarantees numerical identity of parameters after synchronization, it does not eliminate discrepancies in local loss values or gradient vectors that contribute to the aggregated update. Consequently, workers may follow partially divergent optimization paths before re-alignment at each iteration. Under full fine-tuning settings—where all parameters are updated and per-device batch sizes are typically constrained—small cross-worker variations in gradient magnitude or direction can accumulate over time. Importantly, such divergence may not manifest as numerical instability or anomalies in globally aggregated loss curves~\cite{zhang2026training}.

This latent divergence, which we term silent inconsistency, refers to cross-worker misalignment in optimization dynamics that remains invisible under conventional monitoring practices~\cite{ma2025understanding}. While distributed data-parallel training has been extensively optimized from system and scaling perspectives, the majority of prior work implicitly treats synchronization as sufficient for trajectory coherence~\cite{you2019large,narayanan2021efficient,zhang2024does}. This assumption overlooks the possibility that workers may experience transient optimization misalignment at the gradient level, which remains invisible under globally aggregated monitoring metrics. To our knowledge, systematic characterization of such worker-level optimization consistency during full fine-tuning remains largely unexplored. However, this assumption overlooks potential discrepancies in gradient dispersion and directional consistency across workers.

In this work, we focus on monitoring rather than modifying DP training. We propose a lightweight diagnostic framework that quantifies worker-level consistency using training signals already available in standard pipelines. Specifically, we introduce three complementary metrics computed from per-worker losses and gradients: loss dispersion, gradient-norm dispersion, and gradient-direction consistency measured by cosine similarity. These metrics require no changes to model architecture, synchronization mechanisms, or optimization algorithms, and incur negligible computational overhead. Through full fine-tuning of the 1B-parameter Pangu model on an 8-NPU DP setup~\cite{chen2025pangu}, we demonstrate that the proposed metrics reveal divergence patterns that are completely masked by aggregated loss curves and provide actionable insight into the effects of precision settings, data ordering, and learning-rate strategies on training stability.

The main contributions of this work are summarized as follows:

\begin{itemize}
    \item We formalize the phenomenon of \textit{silent inconsistency} in data-parallel full fine-tuning of LLMs, highlighting the distinction between parameter synchronization and optimization trajectory alignment.
    
    \item We propose a lightweight, training-signal-based monitoring framework that quantifies cross-worker consistency without altering existing training pipelines.
    
    \item We empirically demonstrate that conventional aggregated metrics can mask significant worker-level divergence and show how the proposed indicators improve transparency and diagnostic capability in large-scale DP fine-tuning.
\end{itemize}

\section{Related Work}

Parameter-efficient adaptation has become a mainstream strategy for reducing the cost of adapting large language models, exemplified by low-rank updates (LoRA) and continuous prompt optimization (prefix-tuning) \cite{hu2022lora}. Nonetheless, full fine-tuning remains widely used when maximal task-specific performance or broad behavioral reshaping is required, which in turn necessitates scalable distributed training. Within this context, data-parallel (DP) training is often preferred for its architectural simplicity. The DP literature has largely concentrated on system-level scalability and communication efficiency—ranging from synchronization designs to bandwidth-saving techniques—while maintaining convergence guarantees under synchronous aggregation \cite{tang2020communication}. A prevailing practical assumption in this stream is that synchronized parameter updates imply coherent optimization behavior across workers, leaving the pre-aggregation worker-level dynamics insufficiently characterized.

A second research stream concerns reliability and debugging in machine learning systems, where failures may stem from data issues, implementation defects, and pipeline complexity. Prior work has surveyed debugging methodologies across the ML lifecycle, emphasizing that ML failures are frequently silent, non-deterministic, and difficult to localize compared with traditional software faults \cite{nguyen2025systematic}. Related pipeline-oriented studies further highlight how errors propagate across stages of data preparation, training, and evaluation, motivating monitoring mechanisms that can diagnose issues beyond simple performance metrics \cite{karlavs2025navigating}. More recently, proactive checking frameworks such as \textsc{TrainCheck} have proposed automatically inferred invariants to flag silent training errors without requiring explicit crashes or obvious anomalies in global loss curves \cite{jiang2025training}. While these approaches substantially improve practical debuggability, they are primarily designed to detect violations of expected behavior indicative of bugs or misconfigurations, rather than to quantify systematic cross-worker misalignment that can arise even when the training pipeline is functioning ``normally.''

A third line of work is closely related but targets silent data corruption (SDC) originating from hardware faults in large-scale training. Empirical evidence suggests that SDC can perturb intermediate computations and gradients, potentially steering training toward different optima even when conventional aggregated signals appear benign \cite{ma-etal-2025-understanding}. In response, lightweight detection mechanisms have been proposed to identify and localize corrupted devices by introducing checks around collective communication and analyzing gradient-related statistics, aiming to prevent faulty tensors from contaminating global updates \cite{li2026lightweight}. However, SDC-focused methods are oriented toward discrete fault events and device-level corruption. They are not intended to characterize the broader and more routine sources of worker-to-worker divergence—such as data partitioning effects, floating-point non-associativity, mixed-precision arithmetic, and runtime variability—that may occur without any underlying hardware failure.

In summary, prior studies provide (i) scalable DP optimizations that typically treat synchronization as a proxy for trajectory coherence \cite{tang2020communication}, (ii) debugging frameworks that emphasize detecting silent errors via inferred invariants \cite{jiang2025training} or pipeline-level monitoring \cite{karlavs2025navigating}, and (iii) SDC analyses and detectors targeting hardware-induced corruption \cite{ma-etal-2025-understanding,li2026lightweight}. Collectively, these strands leave a practical gap: an online, lightweight, model-agnostic approach for measuring worker-level optimization alignment during synchronous DP full fine-tuning, even when training appears stable under aggregated monitoring. Our work addresses this gap by introducing complementary metrics—loss dispersion, gradient-norm dispersion, and gradient-direction consistency—that directly quantify cross-worker agreement using standard training signals, thereby enabling systematic diagnosis of ``silent inconsistency'' that may otherwise remain hidden.

\section{Methodology}

In this section, we define a set of online monitoring metrics for quantifying worker-level consistency in data-parallel (DP) full fine-tuning. The objective is diagnostic rather than algorithmic: we do not alter synchronization, optimization, or model architecture. Instead, we formalize consistency indicators using training signals already produced during standard forward/backward computation, so the metrics can be integrated into existing DP pipelines with negligible additional cost.

We consider a DP configuration with $N$ workers. At training step $t$, worker $i \in \{1,\ldots,N\}$ processes its local mini-batch and computes a scalar loss $\ell_i^{(t)}$ and a gradient vector $\mathbf{g}_i^{(t)} \in \mathbb{R}^d$ with respect to the trainable parameters (before any cross-worker reduction). All metrics below are defined across workers at the same step and can be logged either before or after gradient synchronization, depending on what signals are accessible in the implementation. Unless otherwise specified, we use the pre-reduction quantities to capture discrepancies introduced prior to all-reduce.

\subsection{ Loss dispersion}
Loss monitoring in practice typically reports a single scalar per step (often a mean loss or the loss from a designated worker), which can mask worker-to-worker variation induced by data sharding and numerical effects. We therefore quantify cross-worker variability of the per-step loss by measuring its dispersion across workers. Let
\begin{equation}
\bar{\ell}^{(t)}=\frac{1}{N}\sum_{i=1}^{N}\ell_i^{(t)}
\end{equation}
denote the mean loss at step $t$. We define the loss dispersion as the standard deviation
\begin{equation}
D_{\mathrm{loss}}^{(t)}=\sqrt{\frac{1}{N}\sum_{i=1}^{N}\left(\ell_i^{(t)}-\bar{\ell}^{(t)}\right)^2}.
\end{equation}
For robustness to outliers, one may additionally report the range
\begin{equation}
R_{\mathrm{loss}}^{(t)}=\max_i \ell_i^{(t)}-\min_i \ell_i^{(t)}.
\end{equation}

Under stable execution with consistent worker behavior, $D_{\mathrm{loss}}^{(t)}$ remains small and fluctuates around a stationary level determined by data heterogeneity and batch sampling noise. Persistent elevation or abrupt spikes in dispersion indicate that some workers are experiencing systematically different training signals, even when the aggregated loss curve remains smooth.

\subsection{ Gradient-norm dispersion}

Loss-level agreement does not imply that workers apply comparable update magnitudes. In mixed-precision training and in regimes with small per-device batch sizes, gradient scaling, rounding, or occasional extreme samples can yield substantially different gradient magnitudes across workers, which may not immediately manifest as loss divergence. To characterize magnitude-level disagreement, we monitor the dispersion of per-worker gradient norms computed prior to gradient aggregation. Let $\mathbf{g}_i^{(t)}$ be the local gradient vector at step $t$. Its Euclidean norm is
\begin{equation}
\left\|\mathbf{g}_i^{(t)}\right\|_2=\sqrt{\sum_{k=1}^{d}\left(g_{i,k}^{(t)}\right)^2}.
\end{equation}
Define the mean gradient norm
\begin{equation}
\bar{g}^{(t)}=\frac{1}{N}\sum_{i=1}^{N}\left\|\mathbf{g}_i^{(t)}\right\|_2,
\end{equation}
and the gradient-norm dispersion
\begin{equation}
D_{\mathrm{grad}}^{(t)}=\sqrt{\frac{1}{N}\sum_{i=1}^{N}\left(\left\|\mathbf{g}_i^{(t)}\right\|_2-\bar{g}^{(t)}\right)^2}.
\end{equation}

Large values of $D_{\mathrm{grad}}^{(t)}$ indicate that workers produce gradients of disparate magnitudes, which can be symptomatic of numerical sensitivity, unstable examples, or runtime perturbations. Tracking this quantity over time provides an early signal of abnormal update behavior that may precede overt numerical failures (e.g., overflow or NaNs).

\subsection{Gradient-direction consistency}

Magnitude-level agreement still does not guarantee that workers agree on the update direction. Directional disagreement is particularly relevant for DP training because all-reduce averages gradients: when local gradients are misaligned, aggregation can partially cancel updates and degrade optimization efficiency. We quantify directional alignment using cosine similarity between local gradient vectors. For a pair of workers $i$ and $j$, define the cosine similarity at step $t$ as
\begin{equation}
\cos\!\left(\mathbf{g}_i^{(t)},\mathbf{g}_j^{(t)}\right)
=\frac{\mathbf{g}_i^{(t)}\cdot \mathbf{g}_j^{(t)}}{\left\|\mathbf{g}_i^{(t)}\right\|_2\,\left\|\mathbf{g}_j^{(t)}\right\|_2}.
\end{equation}
We aggregate pairwise alignment into a single directional-consistency metric by averaging over all unordered worker pairs:
\begin{equation}
C_{\mathrm{dir}}^{(t)}=\frac{2}{N(N-1)}\sum_{1\le i<j\le N}\cos\!\left(\mathbf{g}_i^{(t)},\mathbf{g}_j^{(t)}\right).
\end{equation}

By construction, $C_{\mathrm{dir}}^{(t)}\in[-1,1]$, with values closer to $1$ indicating stronger cross-worker alignment. In practice, reductions in $C_{\mathrm{dir}}^{(t)}$ often provide a sensitive indicator of emerging inconsistency, including cases where losses remain similar across workers but gradients begin to disagree in direction.

\section{Experiments}

\subsection{Dataset and Experimental Setup}

All experiments adopt a \emph{full-parameter fine-tuning} strategy on the Ascend Tribe \texttt{openPangu-Embedded-1B-V1.1} model. We employ the \texttt{tatsu-lab/alpaca} instruction-following dataset and transform each instance into an autoregressive sequence following the unified ``Instruction–Input–Response'' template. The concatenated text is tokenized and truncated or padded to a maximum sequence length of 1024 tokens. To ensure that supervision is concentrated on response generation, loss is computed only over the Response segment, while tokens corresponding to the prompt and instruction template are masked out from optimization.

Training is conducted on a single node equipped with eight Ascend NPUs, each providing 64\,GB of device memory. We utilize data-parallel distributed training based on \texttt{torch.distributed} with \texttt{DistributedDataParallel} (DDP), where each process is bound to one NPU and gradients are synchronized via all-reduce. The dataset is partitioned across ranks using a distributed sampler to satisfy the data-parallel assumption. Mixed-precision training (bf16) is enabled to improve computational efficiency and memory utilization. Apart from randomness control, all experiments share identical model architecture, hardware configuration, and optimization pipeline.

To systematically investigate the effect of data-order and sharding consistency in distributed full fine-tuning, we design three controlled settings. In \textbf{S1-1 (strict consistency)}, all ranks share identical random seeds and deterministic data shuffling through \texttt{DistributedSampler}, ensuring synchronized sample ordering and shard assignment across processes. In \textbf{S1-2 (mild inconsistency)}, only rank~0 is assigned a different random seed while the remaining ranks retain the baseline seed, introducing a limited perturbation to stochastic states. In \textbf{S1-3 (significant inconsistency)}, each rank is assigned a distinct, rank-dependent seed, explicitly breaking cross-rank alignment in sample ordering and local stochastic behavior. These configurations form a progressive spectrum from fully synchronized to deliberately desynchronized distributed training.

To quantify the impact of consistency violations on optimization dynamics, we record the local training loss of each rank together with the gradient norm captured \emph{prior} to gradient all-reduce. Specifically, a DDP communication hook is registered to intercept gradient buckets and accumulate their pre-allreduce $\ell_2$ norms, enabling direct observation of local gradient magnitudes before synchronization. At synchronized steps, reduced global statistics are further computed across ranks and aggregated on rank~0 for analysis. This instrumentation preserves the standard training workflow while providing fine-grained measurements of cross-rank divergence under controlled perturbations of data-order consistency.

\subsection{Results and Analysis}

\begin{figure}[H]
    \centering
    \includegraphics[width=0.9\textwidth]{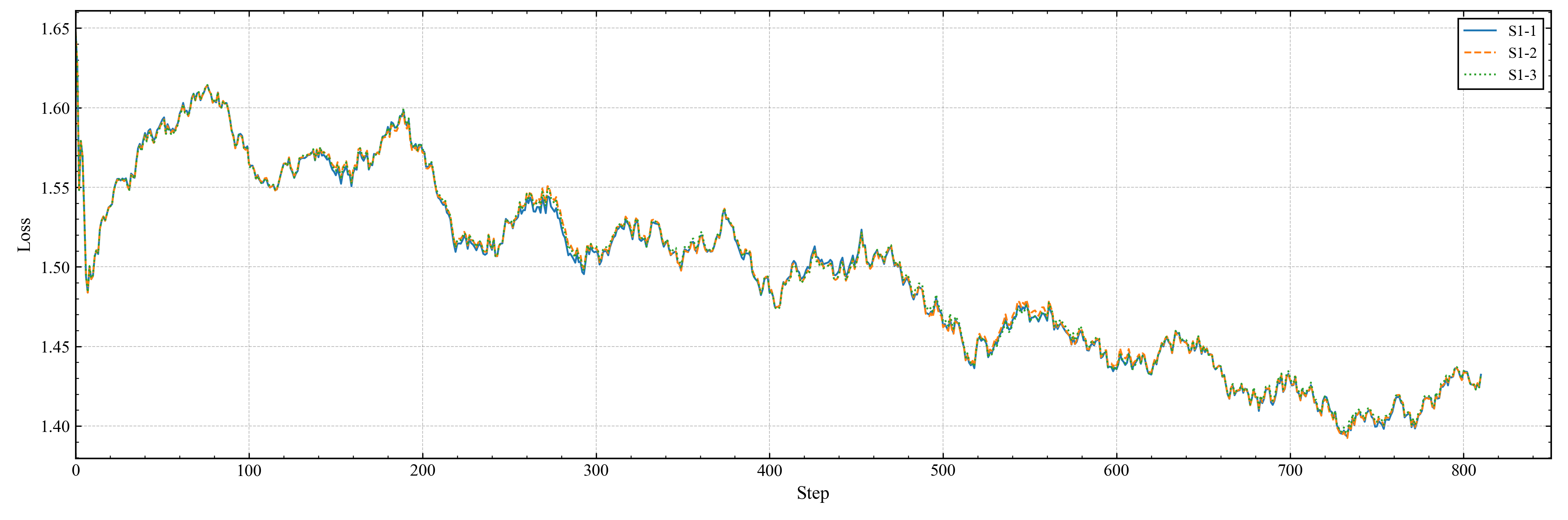} % Replace with your file path
    \caption{Training Loss Curve for Experiments S1-1, S1-2, and S1-3}
    \label{fig:loss_curve}
\end{figure}

\begin{figure}[H]
    \centering
    \includegraphics[width=0.9\textwidth]{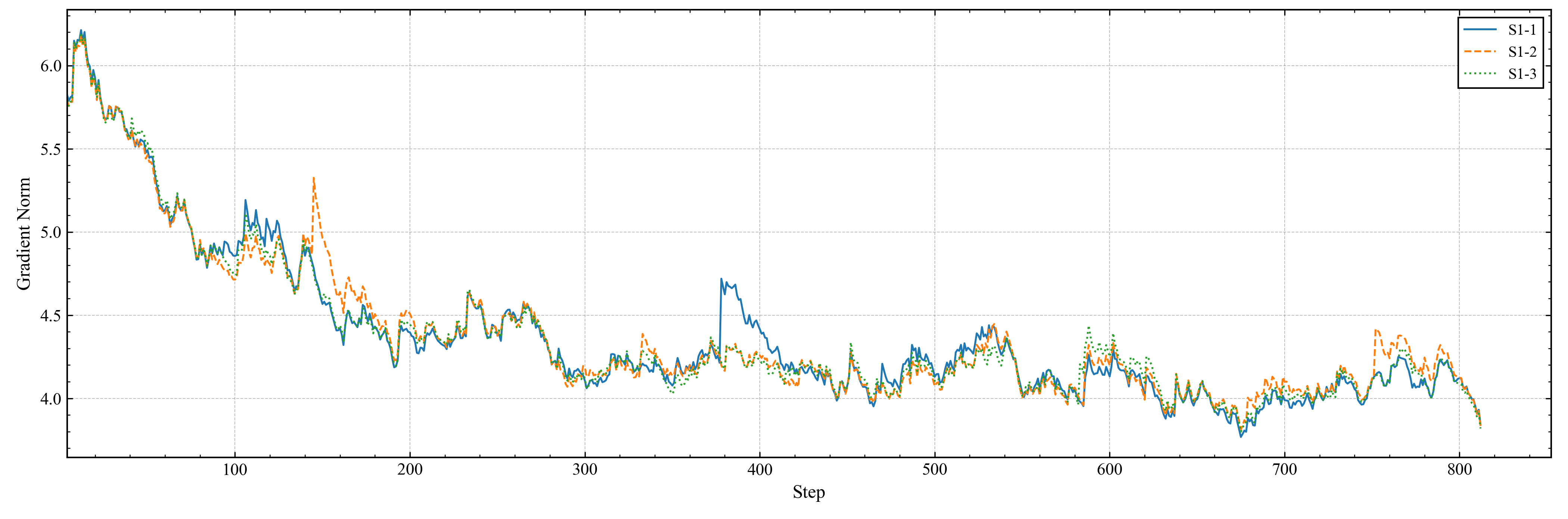} % Replace with your file path
    \caption{Gradient Norm Fluctuations Across Different Experiments}
    \label{fig:gradient_norm}
\end{figure}

The analysis of training loss and gradient norms across the experiments (S1-1, S1-2, and S1-3) underscores the critical role of synchronization in distributed optimization. As illustrated in \figurename~\ref{fig:loss_curve}, experiment S1-1 (Strict Consistency) exhibits a smooth and steady decrease in training loss, indicating stable optimization driven by synchronized random seeds and data shuffling across workers. In contrast, as the degree of inconsistency increases in S1-2 (Mild Inconsistency) and S1-3 (Significant Inconsistency), larger deviations in the loss curves are observed, reflecting the growing impact of stochastic discrepancies between worker behaviors.

Similarly, the gradient norm data, presented in \figurename~\ref{fig:gradient_norm}, highlights the varying levels of optimization stability. For S1-1, the gradient norms remain relatively consistent, suggesting well-aligned optimization trajectories across workers. However, in S1-2, slight variations in gradient magnitudes are observed due to minor inconsistencies in random seed synchronization, while S1-3 reveals more pronounced fluctuations in gradient norms. These fluctuations indicate significant divergence in the optimization paths, exacerbating the inefficiencies of the training process.

\begin{figure}[!t]
    \centering
    \begin{minipage}{0.45\textwidth}
        \centering
        \includegraphics[width=\textwidth]{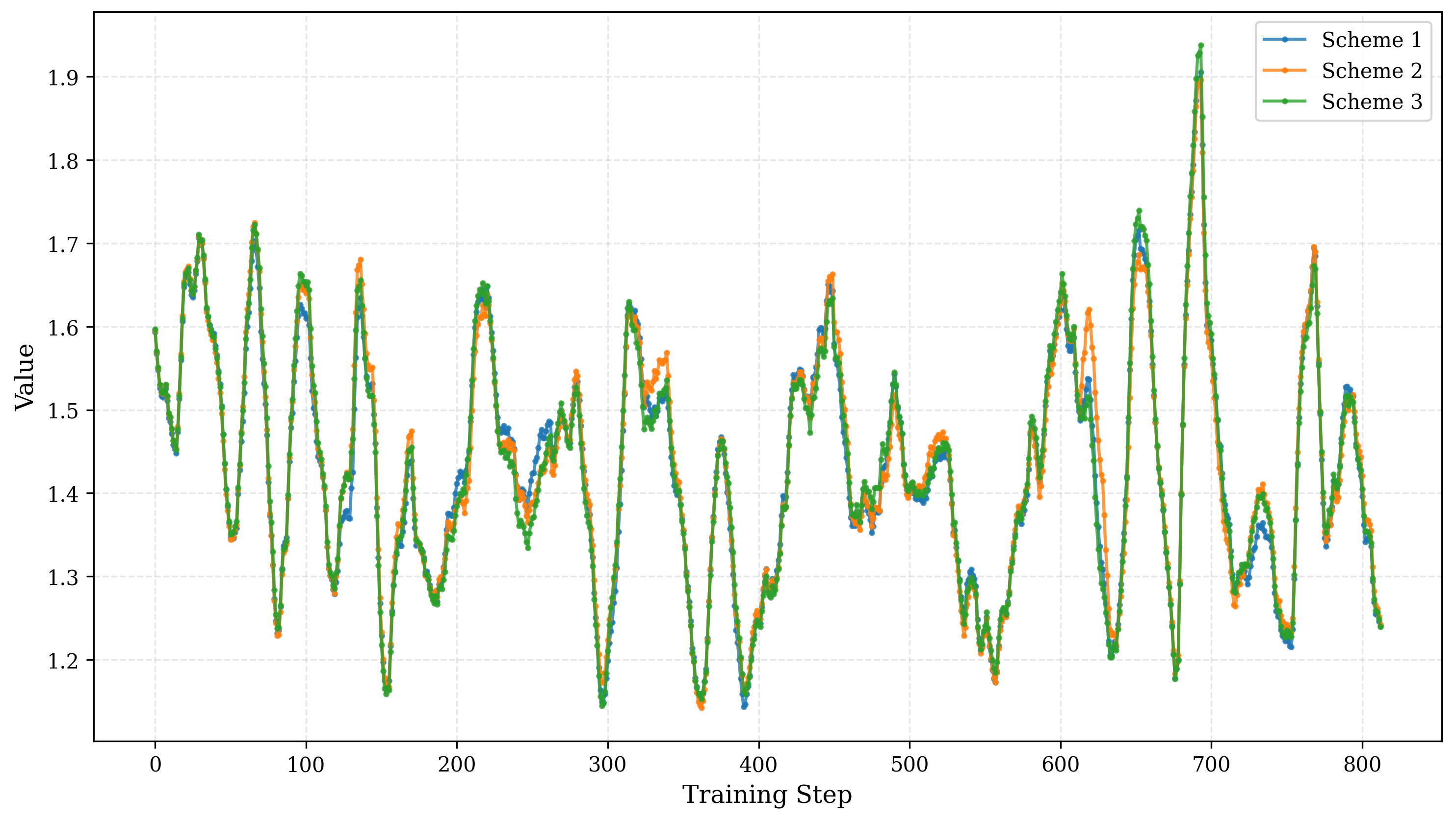} % Adjust the path as necessary
        \caption{Loss Range (Max-Min) for S1-1, S1-2, and S1-3}
        \label{fig:loss_range}
    \end{minipage}%
    \hfill
    \begin{minipage}{0.45\textwidth}
        \centering
        \includegraphics[width=\textwidth]{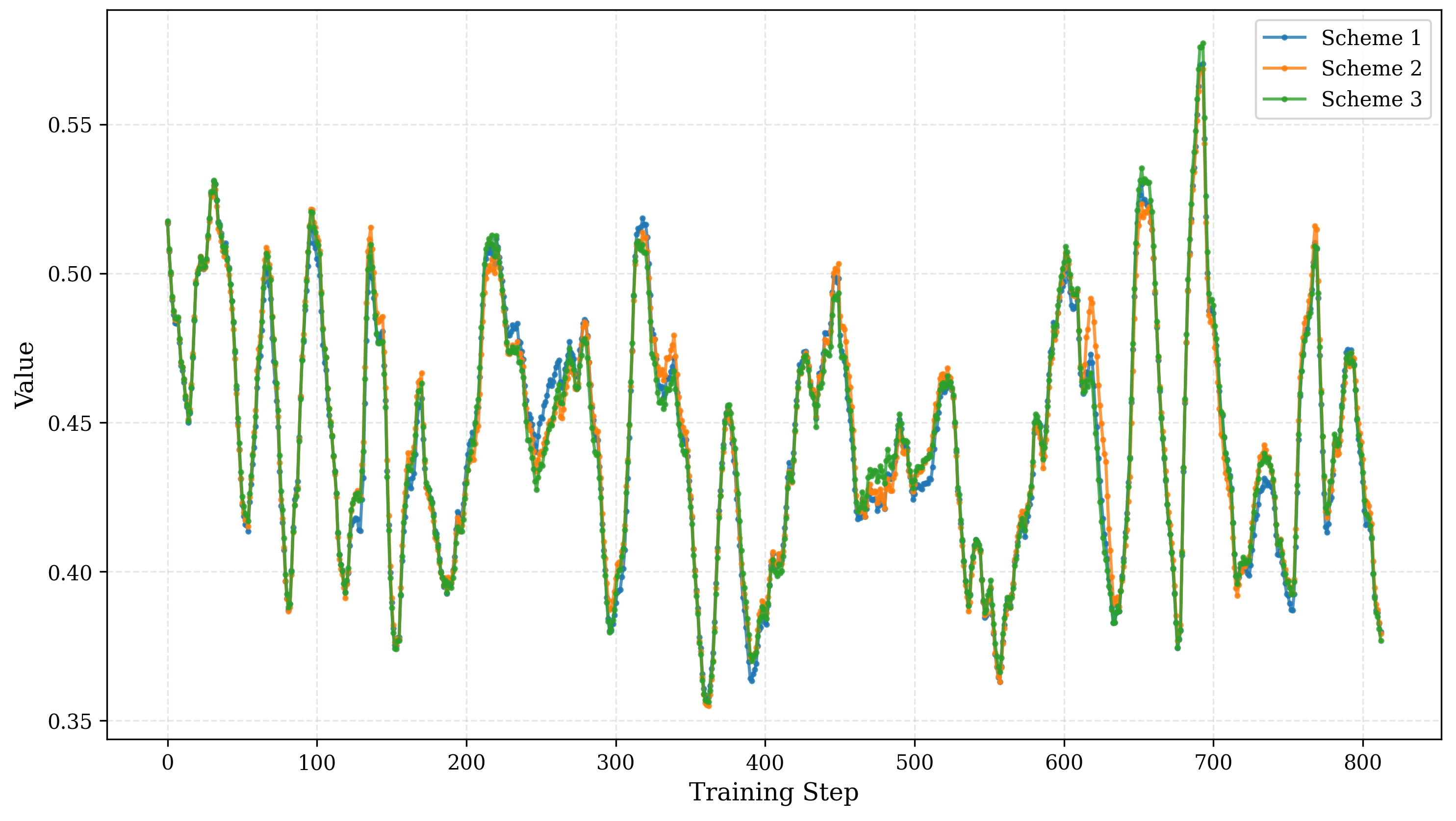} % Adjust the path as necessary
        \caption{Loss Standard Deviation for S1-1, S1-2, and S1-3}
        \label{fig:loss_std}
    \end{minipage}

    \vskip\baselineskip
    
    \begin{minipage}{0.45\textwidth}
        \centering
        \includegraphics[width=\textwidth]{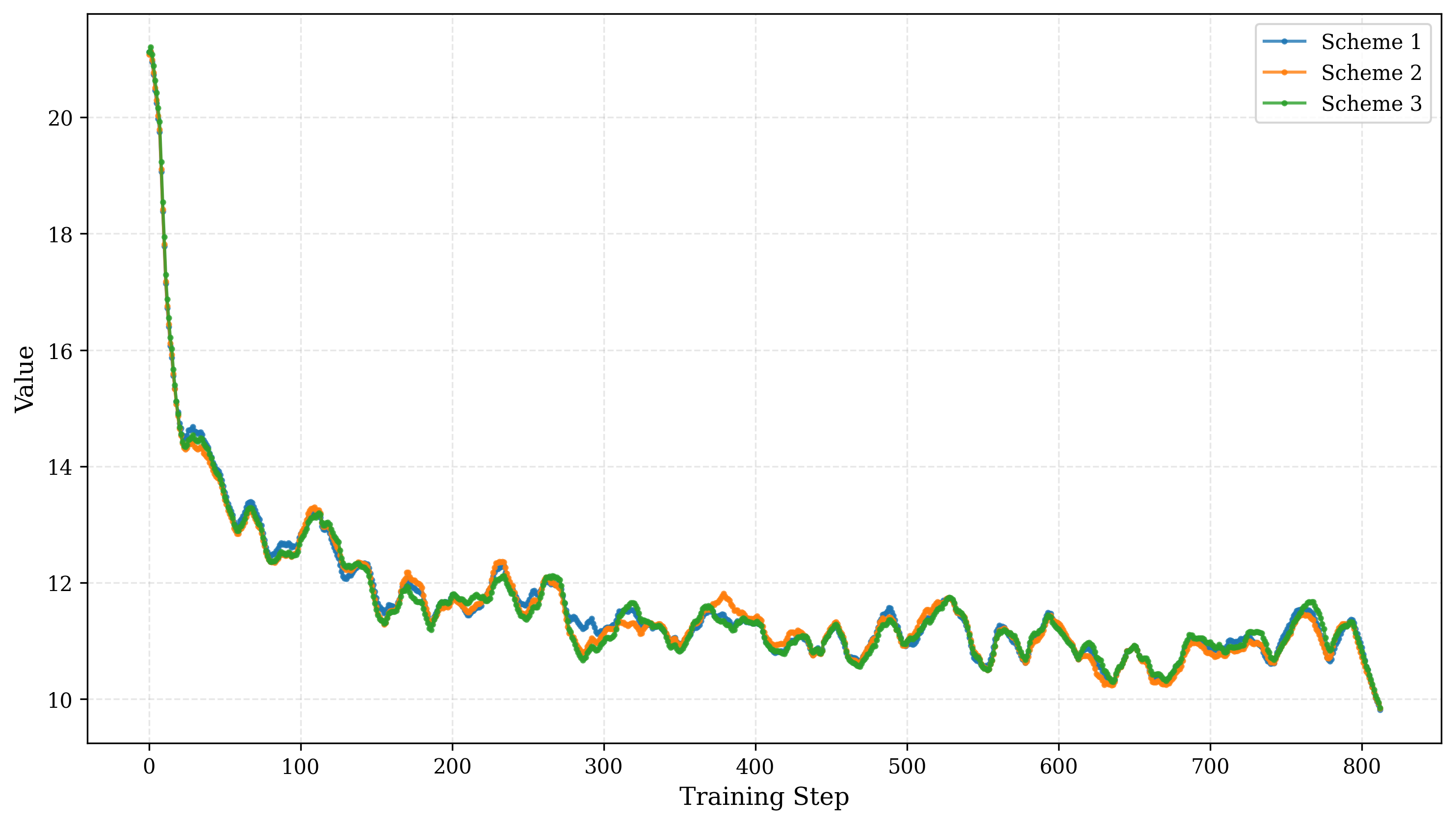} % Adjust the path as necessary
        \caption{Gradient Norm Average for S1-1, S1-2, and S1-3}
        \label{fig:grad_norm_avg}
    \end{minipage}%
    \hfill
    \begin{minipage}{0.45\textwidth}
        \centering
        \includegraphics[width=\textwidth]{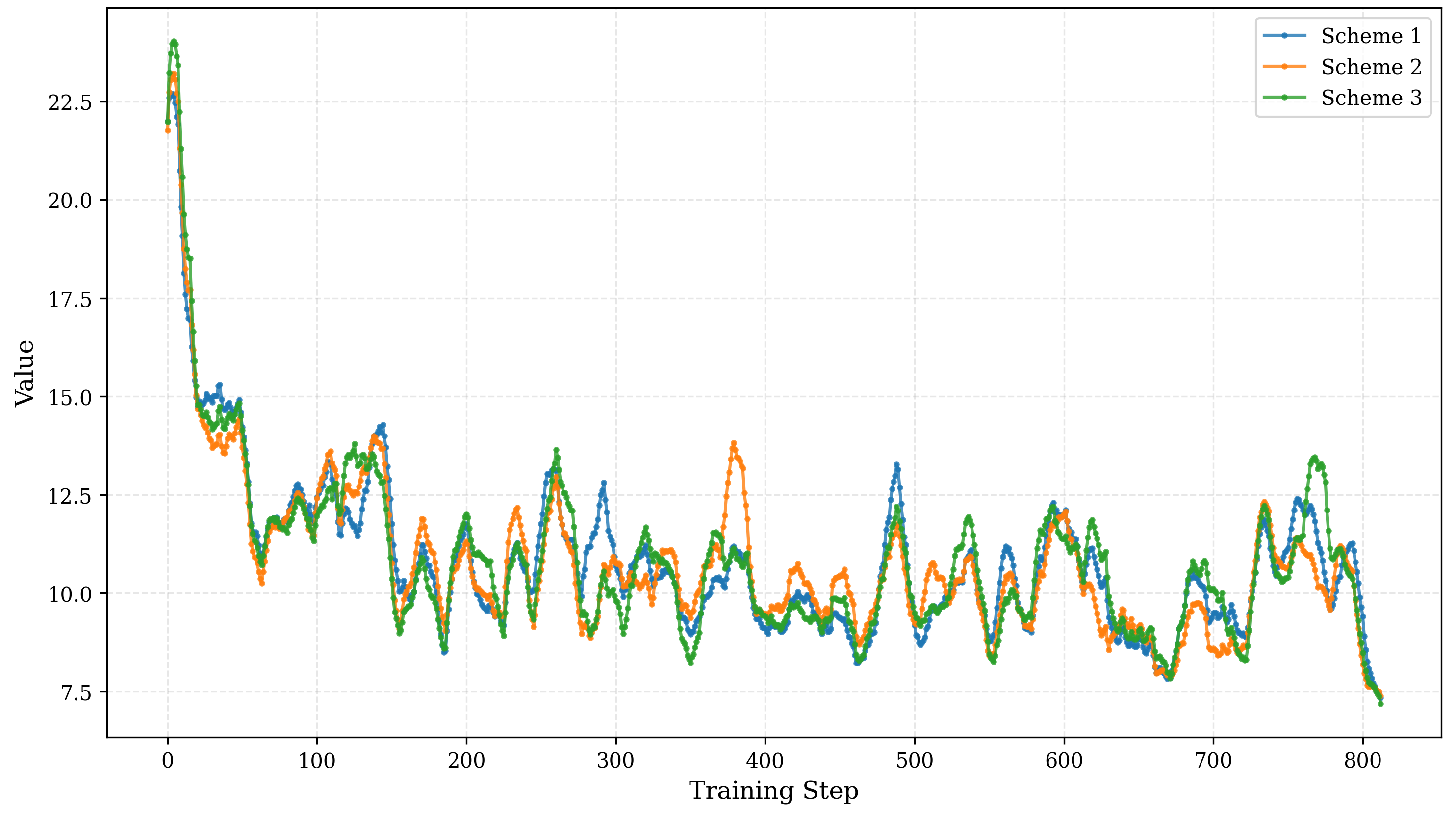} % Adjust the path as necessary
        \caption{Gradient Norm Range for S1-1, S1-2, and S1-3}
        \label{fig:grad_norm_range}
    \end{minipage}

    \vskip\baselineskip
    
    \begin{minipage}{0.45\textwidth}
        \centering
        \includegraphics[width=\textwidth]{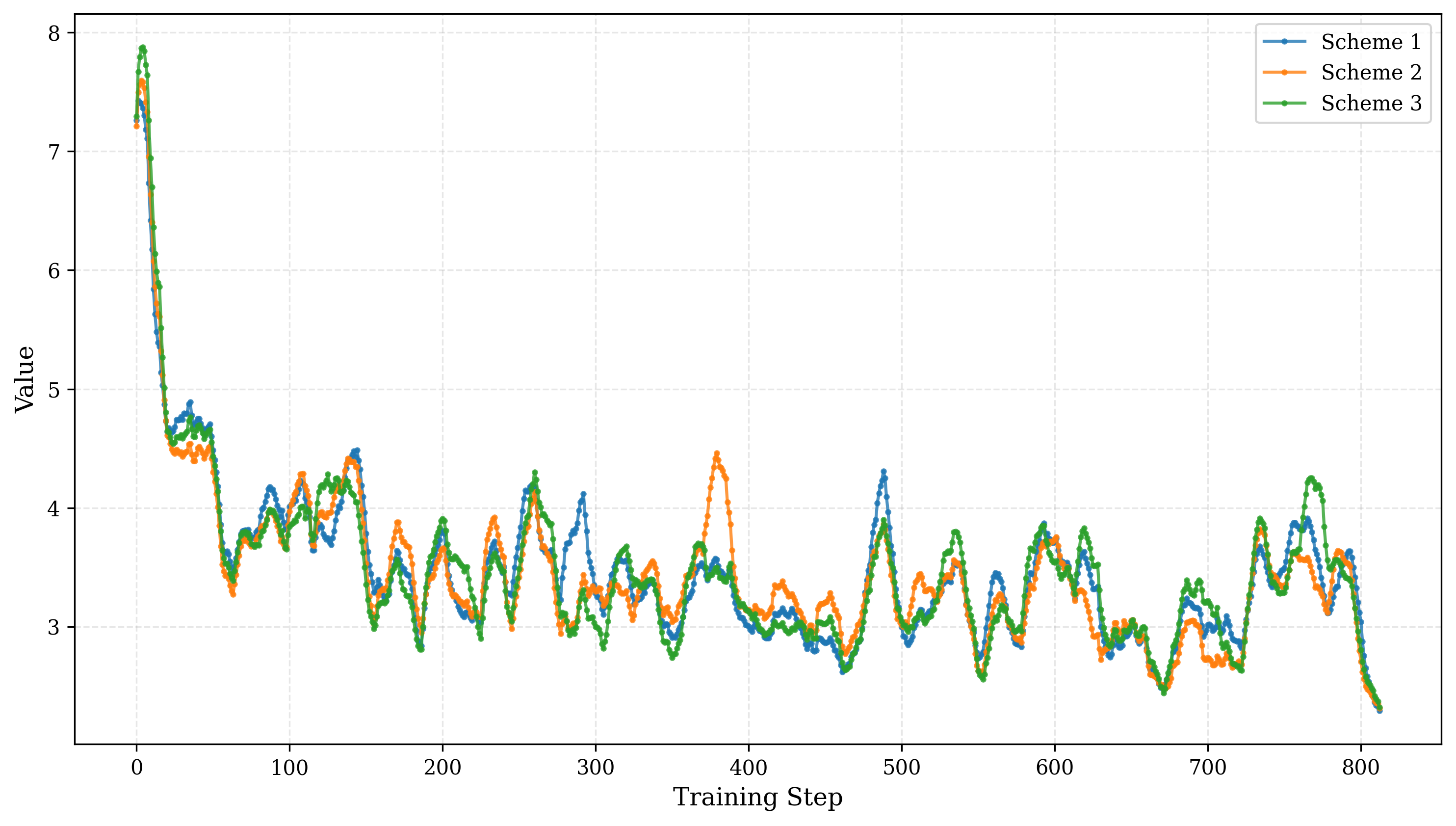} % Adjust the path as necessary
        \caption{Gradient Norm Standard Deviation for S1-1, S1-2, and S1-3}
        \label{fig:grad_norm_std}
    \end{minipage}%
    \hfill
    \begin{minipage}{0.45\textwidth}
        \centering
        \includegraphics[width=\textwidth]{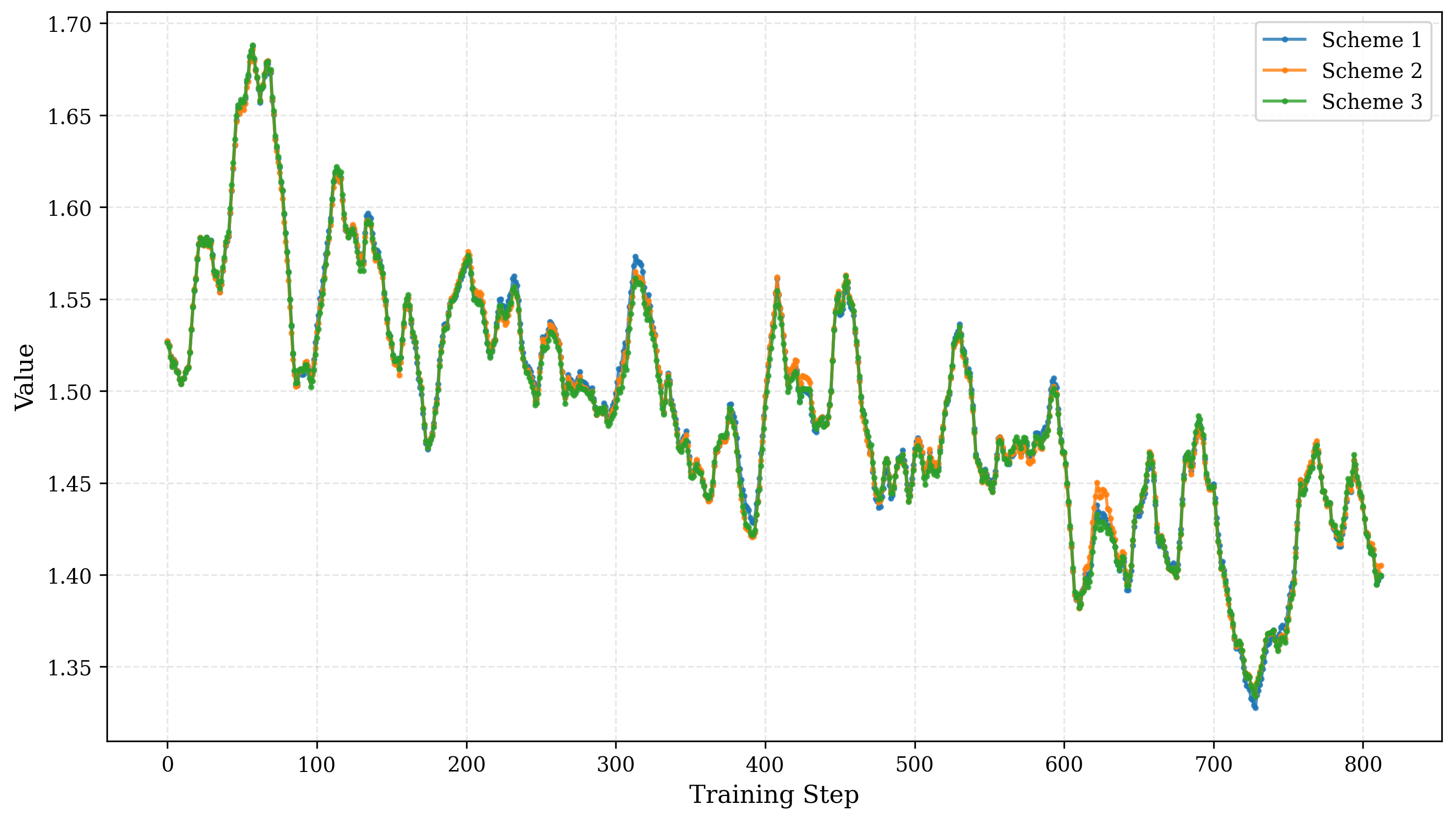} % Adjust the path as necessary
        \caption{Loss Average Across All Ranks for S1-1, S1-2, and S1-3}
        \label{fig:loss_avg}
    \end{minipage}
\end{figure}

Further analysis of additional metrics reveals a more nuanced view of training stability. As shown in \figurename~\ref{fig:loss_range} and \figurename~\ref{fig:loss_std}, the loss range (max-min) and standard deviation metrics provide deeper insight into worker-level discrepancies. In S1-1, the loss range remains tight, and the standard deviation is relatively low, indicating well-coordinated optimization steps. In contrast, S1-2 and S1-3 show notable increases in both loss range and standard deviation, especially in S1-3, where the optimization instability is most evident. These metrics are crucial for understanding not just the average loss but also the variability and spread across workers, which can often remain undetected in aggregated loss values.

Additionally, the gradient norm statistics further validate these observations. As illustrated in \figurename~\ref{fig:grad_norm_avg} and \figurename~\ref{fig:grad_norm_range}, both the average and range of gradient norms are most stable in S1-1, with minimal fluctuations across the training steps. In S1-2 and S1-3, there is a visible increase in both the gradient norm average and range, further indicating that the divergence in worker optimization trajectories is amplified under higher levels of inconsistency.

Finally, gradient norm standard deviation, shown in \figurename~\ref{fig:grad_norm_std}, captures the magnitude of these fluctuations, with S1-1 maintaining the most consistent gradient updates. S1-2 and S1-3 exhibit increasing divergence, particularly in the later stages of training, where higher inconsistencies in the training process lead to broader gradient discrepancies across workers.

Taken together, these results emphasize that maintaining tight synchronization between workers is essential for achieving stable and efficient optimization in distributed training settings. Although minor inconsistencies may not drastically affect overall performance, substantial discrepancies in random seed alignment and data shuffling lead to misalignment in worker behavior, impairing convergence and ultimately resulting in suboptimal performance.

\section{Conclusion}

In this work, we investigated the often-overlooked phenomenon of \emph{silent inconsistency} in data-parallel (DP) full fine-tuning of large language models. Although synchronous all-reduce guarantees parameter equivalence after each iteration, it does not ensure alignment of worker-level optimization dynamics prior to gradient aggregation. We formalized this distinction and introduced a lightweight, model-agnostic monitoring framework that quantifies cross-worker consistency through three complementary metrics: loss dispersion, gradient-norm dispersion, and gradient-direction consistency.

Through controlled experiments on the 1B-parameter Pangu model under progressively desynchronized training configurations, we demonstrated that conventional aggregated metrics—such as globally averaged loss—can conceal substantial worker-level divergence. In contrast, the proposed indicators provide fine-grained visibility into optimization behavior, revealing systematic discrepancies in both magnitude and direction of gradients even when global training curves appear stable. Our empirical results confirm that increasing stochastic inconsistency across ranks leads to measurable degradation in alignment, manifested as elevated loss variability and gradient dispersion.

Importantly, the proposed framework operates entirely on signals already produced in standard DP pipelines and requires no modification to model architecture, synchronization mechanisms, or optimization algorithms. This makes it practical for large-scale deployments where intrusive instrumentation is undesirable. By enhancing transparency into distributed optimization dynamics, our approach supports more reliable debugging, stability assessment, and configuration analysis in full-parameter fine-tuning scenarios.

Future work will extend this monitoring paradigm to larger model scales and heterogeneous multi-node environments, as well as explore adaptive control strategies that respond to detected inconsistency in real time. We believe that systematic measurement of worker-level optimization alignment is a necessary step toward improving robustness and trustworthiness in large-scale distributed training.

\section*{Acknowledgements}
This work was partially supported by SEU Kunpeng \& Ascend Center of Cultivation.

\bibliography{references}

\end{document}